%% file: main.tex
\documentclass{article}
\usepackage{geometry}
\usepackage[hidelinks,colorlinks=true,linkcolor=blue,citecolor=blue]{hyperref}

\usepackage{microtype}
\usepackage{graphicx}
\usepackage{subcaption}
\usepackage{booktabs} 

\usepackage{moreenum}

\usepackage{hyperref}
  \usepackage{enumitem}

\usepackage{amsmath}
\usepackage{amssymb}
\usepackage{mathtools}
\usepackage{booktabs}       
\usepackage{amsthm}
\usepackage{wrapfig}
\usepackage{ulem}
\usepackage[capitalize,noabbrev]{cleveref}

\usepackage{amssymb}
\usepackage{pifont}

\usepackage{multirow}
\usepackage{booktabs}
\usepackage[round]{natbib}

\theoremstyle{plain}

\theoremstyle{definition}

\theoremstyle{remark}

\newcommand{\ntrain}{n_\text{train}}
\newcommand{\ntest}{n_\text{test}}

\newcommand{\svocab}{s_\text{vocab}}

\newcommand{\nout}{n_\text{out}}

\newcommand{\dmodel}{d_\text{model}}

\newcommand{\Ntrain}{N_\text{train}}
\newcommand{\Ntest}{N_\text{test}}

\newcommand{\Nfine}{N_\text{fine}}

\usepackage[textsize=tiny]{todonotes}

\title{Length Generalization in Arithmetic Transformers}
\author{\textbf{Samy Jelassi}\\ 
Princeton University
\and 
\textbf{St\'{e}phane d'Ascoli}\\ 
EPFL 
\and \textbf{Carles Domingo-Enrich}\\ 
New York University 
\and \textbf{Yuhuai Wu} \\
Stanford University \\ 
Google Research 
\and \textbf{Yuanzhi Li} \\ 
Carnegie Mellon University \\ Microsoft Research 
\and \textbf{Fran\c{c}ois Charton} 
\\ Meta AI
 }
\begin{document}

\maketitle

\begin{abstract}
  \input{abstract}

\end{abstract}
\input{introduction}

\input{related_work.tex}

\input{setting}

\input{addition}
\input{modular}

\input{multiplication.tex}

\input{discussion.tex}

\bibliography{references}
\bibliographystyle{plainnat}
\pagebreak

\appendix

 \input{appendix.tex}

\end{document}

%% file: abstract.tex
We examine how transformers cope with two challenges: learning basic integer arithmetic, and generalizing to longer sequences than seen during training. We find that relative position embeddings enable length generalization for simple tasks, such as addition: models trained on $5$-digit numbers can perform $15$-digit sums. However, this method fails for multiplication, and we propose \textit{train set priming}: adding a few ($10$ to $50$) long sequences to the training set. We show that priming allows models trained on $5$-digit $\times$ $3$-digit multiplications to generalize to $35\times 3$ examples. We also show that models can be primed for different generalization lengths, and that the priming sample size scales as the logarithm of the training set size. Finally, we discuss potential applications of priming beyond arithmetic.

%% file: introduction.tex
\section{Introduction}

Transformers \citep{vaswani2017attention} achieve remarkable results in domains ranging from Natural Language Processing (NLP) \citep{vaswani2017attention,devlin2018bert}, to computer vision \citep{dosovitskiy2020image}, reinforcement learning \citep{chen2021decision,janner2021offline}, and program synthesis \citep{austin2021program}. Yet, they struggle on simple tasks, such as integer arithmetic \citep{nogueira2021investigating}. Recent, transformer-based, large language models, such as ChatGPT \citep{schulman2022chatgpt}, can perform arithmetic on small integers, but their performance drops steeply as operands become large.
The text corpora used to train language models is partly responsible for this situation. Most of the problems of mathematics featured in these data sets involve small numbers. In fact, large integers, with $15$ digits or more, almost never appear in print. 
The absence of large numbers in the training data limits the mathematical ability of large language models. To mitigate this, language models must be able to extrapolate the small number arithmetic they have learned, to larger integers.

Most prior works on learning arithmetic with transformers \citep{nogueira2021investigating,power2022grokking} consider the in-distribution setting, where numbers in the training and test sets are drawn from the same distribution. Out-of-distribution experiments, and in particular extrapolation to larger numbers, have so far proven disappointing. 

On the other hand, length generalization in transformers has been widely studied. The seminal paper by \citet{shaw2018self} identified the position embedding (PEs)  as the likely culprit  for their inability to generalize. Indeed, the absolute position embeddings (APEs), used in many implementations, mix the representation of a token with the embedding of its position in the sequence, making trained models very susceptible to changes in  sequence lengths. Since then, several papers have proposed to use relative position embeddings (RPEs), that encode the relative distance between tokens  \citep{shaw2018self,huang2018improved,dai2019transformer,huang2020improve}, or to replace position embeddings by weighted attention schemes \citep{raffel2020exploring,su2021roformer,press2021train}. While these changes improved extrapolation in natural language processing (NLP), their impact on arithmetic tasks has been little studied.


Recent work suggests that large language models can generalize to longer sequences for the addition task, thanks to specialized prompt engineering techniques \citep{zhou2022teaching}. However, results for multiplication are limited to short extrapolation lengths ($7$ digits).


In this paper, we study length generalization in transformers for four basic arithmetic tasks: addition, modular addition, multiplication and modular multiplication. We train models on $5$-digit operations, and investigate their ability to generalize to numbers with up to $20$ digits for addition, and $35$ digits for multiplication. We show that the use of relative position embeddings allows for length generalization in the case of addition and some modular operations. For $5$-digit $\times$ $3$-digit multiplication, we show that \textit{train set priming}: adding a 
\textit{tiny} amount of examples (50 out of 5000) from the target distribution, surprisingly allows the model to length generalize to very long operands (i.e.\ $35$-digit $\times$ $3$-digit multiplications). The paper is organized as follows.

\begin{itemize}[  parsep=1pt] 
\item[--] 
\autoref{sec:setting} presents our experimental setup: problems, data generation, encoding, models, training and evaluation.
\item[--] 
\autoref{sec:addition} demonstrates that, on the addition task, encoder-only transformers using relative position embeddings, can length generalize. 
\item[--] 
\autoref{sec:modular} presents our results for modular arithmetic. In some cases, absolute position embedding allow for length generalization.
\item[--] 
\autoref{sec:multiplication} introduces train set priming and shows that it achieves extrapolation to very long multiplications. \item[--] 
\autoref{sec:discussion} discusses the results, highlights a few additional results and proposes some future directions.
\end{itemize}
\textbf{Contributions.} This paper delivers five key messages.
\begin{itemize}[parsep=1pt]
   \item[--] 
\textbf{Relative position embeddings ensure length generation in addition.} Models trained to add $5$-digit numbers can generalize to $20$-digit operands.
   \item[--] 
\textbf{Simple techniques fail for multiplication.} RPE do not allow length generalization. Fine-tuning on long sequences helps generalize, but  requires a lot of samples from the target distribution. Also, it causes catastrophic forgetting. 
   \item[--] 
\textbf{Train set priming enables length generalization.} For multiplication, adding a tiny amount of long sequences to the training set ($50$ out of the $9 \times 10^{34}$ possible $35$-digit numbers) allows generalization to $35$-digit operands. Remarkably, the number of long sequences is much smaller than the one needed for fine-tuning.
   \item[--] 
\textbf{Priming sample size scales as the logarithm of the train set size.} 
   \item[--] 
\textbf{Primed model can extrapolate to several lengths.} A model trained to multiply $5$-digit numbers can be primed, with $500$ priming examples, to generalize to numbers with $6$ to $35$-digits. On the other hand, $500$ examples \textit{along} would be far from sufficient to train a model to multiply $6$ to $35$ digits.
\end{itemize}

\textbf{Remark}: In our multiplication experiments, we arbitrarily fix the second operand to have 3 digits. This is to ensure that the task is challenging enough. Regarding the first operand, we arbitrarily set the extrapolation to 35 in order  to hightlight that our models are really able to do length generalization when using priming. However, we believe that our empirical results would still hold when extrapolating to any reasonable length.

%% file: related_work.tex
\section*{Related work}

\textbf{Transformers for mathematics.} Early applications of transformers to mathematics focus on symbolic computations. \citet{lample2019deep} trained them to perform symbolic integration and solve differential equations. 
 \citet{polu2020generative} applied them to theorem proving, \citet{hahn2020teaching} to  temporal logic, and \citet{dersy22} trained them to simplify formulas involving polylogarithms. 
\citet{nogueira2021investigating} investigates their limitations on basic arithmetic operations. 
\citet{palamas2017investigating} experiments with modular arithmetic, and \citet{wenger2022} demonstrates that universal transformers can be trained to perform modular inversion. 
Despite their limitations in arithmetic, \citet{charton2021linear} shows that transformers can perform numerical calculations, like computing eigenvalues or inverting matrices.\newline

 With the advent of large language models \citep{bommasani2021opportunities}, a new line of research focuses solving problems of mathematics written in natural language \citep{ griffith2021,meng2019,cobbe2021training}. \citet{lewkowycz2022solving} show that a large pre-trained transformer can be retrained on a large math corpus to solve grade and high school problems of mathematics.\newline

\textbf{Length generalization with transformers.} Multiple works observe the difficulty of transformers to length generalize  especially in NLP \citep{shaw2018self,murray2018correcting,rosendahl2019analysis,press2021train}. Several techniques have then been introduced to address this problem: new position embeddings \cite{shaw2018self,dai2019transformer,raffel2020exploring,huang2020improve,kiyono2021shape,su2021roformer,press2021train}, introducing new tokens \cite{newman2020eos}, new attention mechanisms \cite{dubois2019location}. In this paper, we leverage one of these techniques (RPE) for addition and introduce a new one, train set priming, for multiplication.\newline

\textbf{Length generalization in mathematics.} Generalization to long sequences, in arithmetic operations, is a longstanding problem. Using recurrent architectures, \citet{joulin2015inferring} and \citet{kaiser2015neural} achieve length generalization in the case of binary addition and multiplication. Later, \citet{trask2018neural} introduces NALU, an architecture that learns addition and multiplication, and that generalizes to any length. However, their network has hand-crafted modules that are specifically designed to encode addition and multiplication. Several recent works use auto-regressive models to length generalize in math tasks. \citet{anil2022exploring} and \citet{zhou2022teaching} show that fine-tuning or scratchpad  \citep{nye2021show,wei2022chain} on autoregressive decoder models is insufficient to length generalize. They tackle this by changing the scratchpad procedure and designing new prompt engineering techniques.  Closer to our work, \citet{zhang2022unveiling} train encoder-only models to length generalize on variable assignment tasks. 



%% file: setting.tex
\section{Experimental setup}\label{sec:setting}

\subsection{Problems and encodings}

We consider \textbf{four arithmetic tasks}: 
\begin{itemize}[ parsep=1pt]
   \item[--] \textbf{Addition}: $y=x_1+x_2$.
   \item[--] \textbf{Modular addition}: $y\equiv x_1+x_2\; [c]$.
   \item[--] \textbf{Multiplication}: $y=x_1\times x_2.$ 
   \item[--]\textbf{Modular multiplication}: $y\equiv x_1\times x_2\;  [c]$,
\end{itemize}
\begin{figure}[t]
\centering

\includegraphics[width=0.6\linewidth]{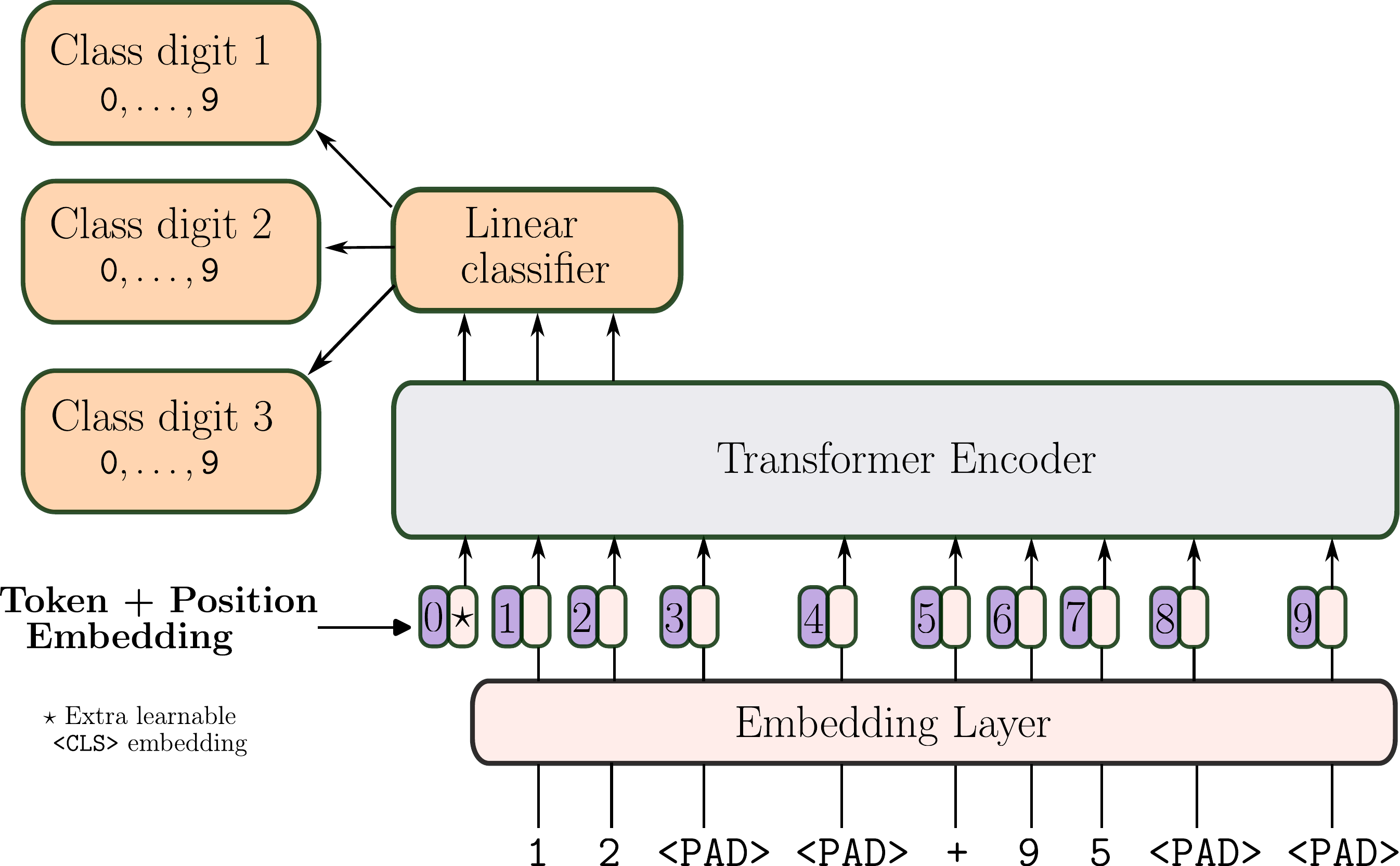}
\caption{Model overview. We  linearly embed each symbol token,
add position embeddings, and feed the resulting sequence of vectors to a transformer or universal transformer
encoder.  In order to predict the result of the operation, we select the first $\nout$ tokens and apply a linear classifier to each of them.}\label{fig:architecture}

\end{figure}

with $x_1$ and $x_2$, two positive integers, and $c > 1$, a fixed modulus. Our models are trained to predict $y$ from ($x_1, x_2)$.

For the addition tasks, the train set is composed of pairs of positive integers with up to $5$ digits, i.e. $(x_1,x_2) \in \mathbb{N}_{10^5}^2$. $x_1$ is randomly sampled from a fixed set of $\Ntrain$ values (we usually set $\Ntrain=5000$). $x_2$ is uniformly sampled in $\mathbb{N}_{10^5}$. Since $\Ntrain \ll 100,000$, the training set only covers a small portion of the problem space. This guarantees that the model will not overfit. Trained models are tested on random pairs of positive integers with $\ntest$ digits: $(x_1, x_2)\in \mathbb{N}_p^2$, $p=10^{\ntest}$. We set $\ntest=5$ for in-domain accuracy, and $\ntest \in \{6,... 20\}$ for length generalization.\newline

For multiplication, we train from pairs of positive integers with up to $5$-digits and $3$-digits, i.e. $x_1<10^5$ and $x_2<10^3$. We henceforth refer to this setting as ``$5\times 3$ multiplication''. As before, $x_1$ is randomly sampled from a fixed set of $\Ntrain$ examples, and $x_2$ is uniformly sampled in $\mathbb{N}_{1000}$. Trained models are tested on $\ntest\times 3$ products, with $\ntest=5$ in-domain, and $\ntest
\in \{6, ... 35\}$ for length generalization. \newline



\textbf{Data formatting.} The arithmetic operations (e.g. $535\times 257$) and the integers ($137495$) that correspond to model input and output are encoded as sequences of discrete symbols. Integers are represented as sequences of digits, in base $10$, and padded (using the special token \texttt{<PAD>}) to lengths $\ntest$ for input operands, and $\nout$ for output. We have $\nout=\ntest+1$ for addition, and $\nout=2 \ntest$ for multiplication. The four operations are encoded with the dedicated tokens \texttt{+}, $\texttt{\%}$, \texttt{$\times$} and \texttt{$*$}. 
Overall, we use a vocabulary of $15$ tokens: $\{\texttt{0},\dots,\texttt{9},\texttt{+},$\texttt{\%}$,\texttt{$\times$},\texttt{$*$}  ,\texttt{<PAD>}\}$. For example, for addition with $\ntrain=2$ and $\ntest=3$, the train and test examples $12+39=51$ and $999+345=1344$ would be encoded as:
 \begin{quote}
 
    $x^{\text{train}} = \texttt{1 2 <PAD> + 3 9 <PAD>}\\
    y^{\text{train}} = \texttt{5 1 <PAD>}\\
    x^{\text{test}} = \texttt{9 9 9 + 3 4 5 }\nonumber\\ 
    y^{\text{test}} = \texttt{1 3 4 4}\nonumber$

\end{quote}

We use the padding symbol in order to ensure that all the input sequences and output sequences have the same length. This is crucial for the model in order to deal with carries.

\textbf{Training procedures.} We use the following three procedures. Standard training is used in Sections~\ref{sec:addition} and~\ref{sec:modular}. Fine-tuning and priming are introduced in \autoref{sec:multiplication}. In all training procedures, the first operands and randomly sampled from a fixed set of $\Ntrain$ examples, and the second operands are generated online (i.e. uniformly sampled between $1$ and $10^5$ for addition, and between $1$ and $10^3$ for multiplication).

\begin{itemize}
  \item[--] \textbf{Standard training}: the model is trained on $\Ntrain$ examples of $\ntrain$-digit integers.  
  \item[--] \textbf{Fine-tuning}: the model is trained on $\Ntrain$  examples of $\ntrain$-digit integers and then fine-tuned on $\Nfine$ examples of $\ntest$-digit integers.
  \item[--] \textbf{Train set priming}:  the model is trained on $(1-\varepsilon)\Ntrain$ examples of $\ntrain$-digit integers and $\varepsilon\Ntrain$ \textit{priming examples} of $\ntest$-digit integers, with $\varepsilon\ll 1$. The priming examples are fixed throughout the training. 
\end{itemize}




\textbf{Evaluation sets.} During and after training, model performance is evaluated on randomly generated test sets, of $\Ntest$ integers with $n$ digits. The resulting accuracy is said to be \textit{in-distribution} (ID) when $n=\ntrain$, and \textit{out-of-distribution} (OOD) when  $n>\ntrain$. New test sets are generated online for each evaluation step.
If not specified otherwise, we use $\ntrain=5$, $\Ntrain=5000$, and $\Ntest=10000$. We set $\ntest=20$ for addition, and $\ntest=35$ for multiplication.




\subsection{Model and training}

\paragraph{Model.} 
We experiment with two encoder-only architectures: a regular transformer \citep{vaswani2017attention}, and a universal transformer (UTransformer) \citep{dehghani2018universal}, in the HuggingFace implementation \citep{wolf2020transformers} of BERT \citep{devlin2018bert} and ALBERT \citep{lan2019albert}.  Our model is a stack of three components (see \autoref{fig:architecture}):

\begin{enumerate} 
    \item \textbf{Embedding}: a  ($\svocab\times\dmodel$)-trainable embedding layer and a position embedding.
    \item \textbf{Encoder}: an  encoder-only transformer or UTransformer. 
    \item \textbf{Classifier}: encoder output is truncated (to its first $\nout$ elements, forming a $\nout\times \dmodel$ matrix), which is processed by a linear layer that outputs $\nout \times \svocab$ predictions, and encodes each symbol as a one-hot vector.
\end{enumerate}

\textbf{Important note:} Although we use the HuggingFace implementation, our encoders are not pre-trained, and we do not use masked language modelling. We train non-causal encoders in a supervised way, using cross-entropy loss.\newline

 

\paragraph{Notes on design.} 
We chose to use universal transformers, i.e.\ transformers with  shared layers \citep{dehghani2018universal}, because recurrent models are used in prior work on length generalization \citep{bansal2022, kaiser2015neural}, and universal transformers proved essential on tasks involving modular arithmetic \citep{wenger2022}. We believe shared-layer architectures are central to solving arithmetic problems, because they embed the recursive nature of many algorithms. They also seem fit for extrapolation tasks where a long operand is processed by successive applications of a simple technique (e.g. one-digit add and carry).\newline


The choice of an encoder-only model contrasts with concurrent works that consider decoder-only  \citep{power2022grokking,bueno2022induced,zhou2022teaching} or sequence to sequence (seq2seq) models \citep{nogueira2021investigating}. We believe that autoregressive models, such as the decoder-only architecture, are not optimal for problems of arithmetic, because they are trained to learn the correlations between successive tokens in the input sequence. 
In natural language, these correlations are meaningful: they represent the syntactic and grammatical relations between words in a sentence. In arithmetic, these correlations are tiny: knowing that the first three digits of number 1234 are 1, 2 and 3, offers no clue about the value of the fourth digit. As for seq2seq models, in problems where output are guaranteed to be shorter than input, we consider an auto-regressive decoder as an unnecessary complication. Overall, we choose encoder-only models because they are the simplest architecture that can address our problems.\newline

\paragraph{Learning problem.}  We frame our arithmetic tasks as the following supervised multi-classification problem:
\begin{align}\label{eq:problem}
    \hspace{-.3cm}\min_{\theta\in\Theta} \sum_{i=1}^{\Ntrain} \sum_{j=1}^{\nout} \sum_{k=1}^{\svocab} \mathbf{1}[y_{i}[j]=k-1]\frac{e^{f_{\theta}(x_i)[j,k]}}{\sum_{k'=1}^{\svocab} e^{f_{\theta}(x_i)[j,k']}},
\end{align}
where $f_{\theta}(x_i)\in\mathbb{R}^{\nout\times\svocab}$ are the model logits evaluated at $x_i$ and $\theta\in\Theta$ are the model parameters. 
To solve \eqref{eq:problem}, we minimize the cross entropy between model predictions and the ground truth symbols for each position in the sequence. An alternative approach, perhaps more natural, would consider these problems as regressions. However, prior works report that reformulating regression as classification leads to state-of-the-art performance \citep{rothe2015dex,rogez2017lcr,akkaya2019solving,schrittwieser2020mastering}.\newline 

We consider three \textit{model sizes}. Base (B) models have $D$=$6$ layers,  $\dmodel$=$512$ dimensions, and $h$=$8$ attention heads, Standard (S) models have $D$=$6$, $\dmodel$=$1024$ and $h$=$16$, and Large (L) models, we have $D$=$10$, $\dmodel$=$1024$ and $h$=$16$. We investigate three kinds of position embeddings: absolute (APE) \cite{vaswani2017attention}, relative over keys (RPE$_k$) \cite{shaw2018self}, and relative over keys and queries (RPE$_{k,q}$) \cite{huang2018improved}. RPE$_k$ is our default option. 
 All other parameters are set to the default HuggingFace values, and are initialized with random Gaussian values. \newline

\paragraph{Optimization.} We train our models using AdamW \citep{loshchilov2017decoupled}, with a batch size to 32, a learning rate between $10^{-5}$ and $10^{-4}$ and weight decays in $\{1\mathrm{e}{-5},1\mathrm{e}{-4},1\mathrm{e}{-3},1\mathrm{e}{-2}\}$. We apply a cosine scheduler \cite{loshchilov2016sgdr} to update the learning rate and train the model for 15000 epochs of $\Ntrain$ examples.


%% file: addition.tex
\section{Addition: relative position embeddings enable length generalization}\label{sec:addition}

\begin{table}[t]

    \small
    \centering
    \begin{tabular}{lllcccc}
        & & & \multicolumn{4}{c}{Number of digits}\\
        Encoder & PE & Size & 6 & 10 & 15 & 20 \\ 
        \midrule
        \multirow{6}{*}{Transformer} & \multirow{2}{*}{APE} & B  & 1.8 & 0 & 0 &0 \\
        & & L &  1.9 & 0 & 0 & 0 \\
        & \multirow{2}{*}{RPE$_k$} & B & 100 &  99.9 & \textbf{97.2} & \textbf{21.3}\\
        & & L &  98.9 & 74.6 & 47.3 & 0.4\\
        & \multirow{2}{*}{RPE$_{k,q}$} & B & 96.8 & 81.1 & 25.0 & 1.1 \\
        && L & 100 & 99.6 & 88.2 & 19.2 \\        
        \midrule
        \multirow{6}{*}{UTransformer} & \multirow{2}{*}{APE} & B & 2.0 & 0 & 0 &0 \\
        & & L & 3.1 & 0 & 0 & 0 \\
        & \multirow{2}{*}{RPE$_k$} & B & 92.1 & 70.6 & 31.2 &0.1 \\ 
        & & L &  100 & 99.9 & \textbf{98.3} & \textbf{18.2}\\ 
        & \multirow{2}{*}{RPE$_{k,q}$} & B & 99.7 & 22.5 & 0 & 0\\
        && L & 90.8 & 58.0 & 31.1 & 1.4\\

       \bottomrule
    \end{tabular}
\caption{\small \textbf{Addition}: Impact of encoder type, size and position embeddings on length generalization. We consider transformers and UTransformers in their Base (B) and Large (L) format, using three position embeddings methods (APE, RPE$_k$, RPE$_{k,q}$). We evaluate different degrees of extrapolation: easy (6 digits), medium (10 digits) and hard (15 and 20 digits). The models are trained on 5000 examples with 1 to 5 digits and we report the accuracy reached by the models on 100,000 example test sets. Results are averaged over 3 seeds.} 

    \label{tab:addition_results}
\end{table}

In these experiments, we train transformers to add two numbers with up to five digits, and test trained models on sums of numbers with $6$ to $20$ digits. We compare the Transformer and UTransformer encoders, in their Base (6 layers, 512 dimensions, 8 attentions heads) and Large (10 layers, 1024 dimensions, 16 heads) configurations, using three position embeddings: absolute, relative on keys, and relative on keys and queries. All models achieve $100\%$ in-domain accuracy. We make the following observations (\autoref{tab:addition_results}):
\begin{itemize} 
   \item[--]
\textbf{Models using the absolute position embedding fail to generalize}. Our best models achieve $3.1\%$ accuracy on 6-digit test examples, and $0\%$ for all longer lengths. This was observed in previous works \cite{shaw2018self,dai2019transformer,huang2020improve,kiyono2021shape}.
   \item[--]
\textbf{Models using relative position embedding generalize to longer sequences.} Our best models achieve $99.9\%$ accuracy on $10$-digits test sets, and $98.3\%$ on $15$-digit sets. Performance drops for longer sequences: we achieve $21.3\%$ for $20$-digits numbers. We remark that the RPE key variant is crucial for achieving extrapolation.
\end{itemize}
 
In APE models, because the position embedding is added to the embedding of every token, the rules of addition must be learned separately for every position. At test time, a model trained on operands with $5$ digits only will not know how to handle digits in position $6$, or $7$, even though it has learned to add digits in position $1$ to $5$. 
Further discussion of the role of position embeddings, and additional experiments on model failures, can be found in \autoref{sec:discussion}.\newline


\begin{figure}[t]

\begin{subfigure}{0.45\columnwidth}
\includegraphics[width=.98\linewidth]{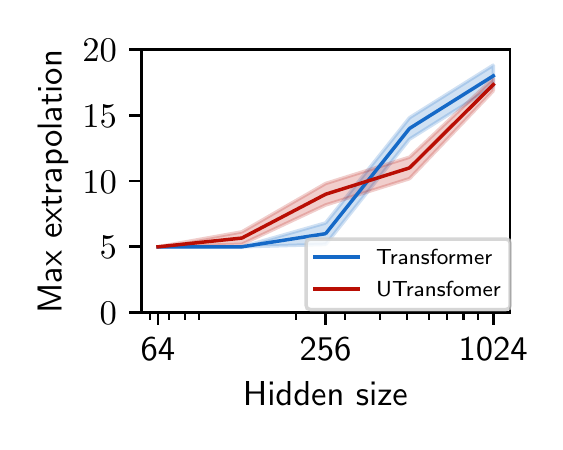}
\vspace*{-7mm}
\caption{}\label{fig:hiddensize}
\end{subfigure}
\begin{subfigure}{0.45\columnwidth}
\includegraphics[width=.94\linewidth]{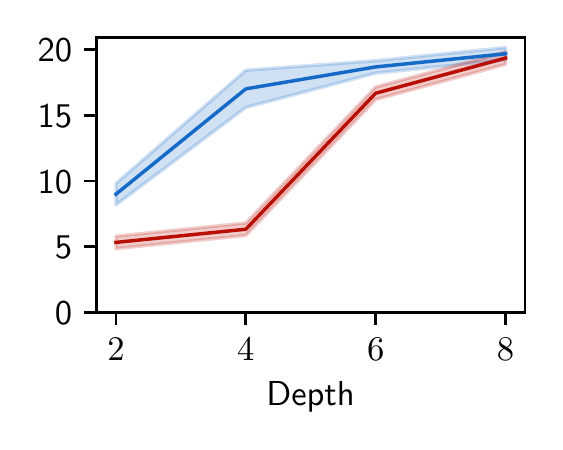}
\vspace*{-7mm}
\caption{}\label{fig:depth}
\end{subfigure}
\vspace{-3mm}
\caption{\small \textbf{Scaling laws for integer addition}.\ We train Transformers and UTransformers, with standard model size ($\dmodel$=$16,D$=$6,h$=$16$) to add numbers with up to $5$ digits. We set $\Ntrain=50000$. We vary their hidden size (a) and depth (b). The $y$-axis indicates the largest extrapolation length where the model achieves 75\% accuracy. Results are averaged over 3 seeds.
}


\end{figure}

\textbf{Depth and dimension for longer extrapolation.} Figures~\ref{fig:hiddensize} and~\ref{fig:depth} provide ablation results on model dimension and depth. For models with $64$ to $1024$ dimensions and $2$ to $8$ layers, trained on $5$ digit examples, they indicate the largest extrapolation length that the model can achieve with $75\%$ accuracy. A minimal hidden size of $512$ for Transformers, and $256$ for UTransformers, is needed for the model to extrapolate. Past this value, length extrapolation scales with dimension, and $1024$-dimension models achieve $17$-digit extrapolation. UTransformers need $6$ layers to extrapolate, whereas shallow Transformers with $2$ layers can extrapolate to $10$-digit numbers. The efficiency of shallow transformer models for computational tasks was observed in previous works \citep{charton2021linear}.





%% file: modular.tex
\section{Modular arithmetic}\label{sec:modular}

In this section, we study modular addition $y\equiv (x_1 + x_2)\;   [c]$ and multiplication $y\equiv (x_1\times x_2)\; [c]$, for $c \in \{100,101,128,1000\}$. The difficulty of these operations depends on the modulus $c$. When $c$ is a power of $10$, i.e. $c=10^k$, modular operations \textit{only} involve the $k$ last digits of their operands, and the result has constant length $k$.  
This makes these operations easier to learn (because they only involve $k$ digits), and easier to generalize (because $k$ is independent of the length of the operands). When the modulus is not a power of $10$, the problem becomes harder than tbeir non-modular verison, because modularity adds an integer division on top of the operation (addition or multiplication).\newline


\textbf{Modular addition.} In the ``easy'' cases ($c\in\{100,1000\}$), RPE-based models generalize to large numbers, achieving better extrapolation performance than for non-modular addition (\autoref{tab:modular_addition}). This is expected, because this is an easier task than standard addition.
Interestingly, APE-based models do generalize; they achieve $73.3\%$ accuracy on $10$-digit numbers. This confirms our intuition that the failure of APE on length generalization is a consequence of their inability to deal with change in output sequence lengths.\newline

For the hard cases ($c\in \{101, 128\}$), no model manages to learn $5$-digit modular addition in-domain. Scaling to larger architectures, with up to $14$ layers and $1280$ dimensions, brings no improvement. This matches previous observations by \citet{palamas2017investigating}, about the difficulty of learning modular arithmetic in the general case.  
\begin{table}[t]
 \begin{subtable}{.5\textwidth}
\small
 \begin{tabular}{clccccc}
        & & \multicolumn{5}{c}{Digits}\\
        Modulo & PE & 5 & 6& 10 & 15 & 20 \\ 
        \midrule
        \multirow{3}{*}{100} & APE & 100 & 99.5 & 73.3 & 43.4 & 21.3  \\
        & RPE$_k$ & 100 & 100 & 86.9 & 36.0 & 3.4\\
        & RPE$_{k,q}$ & 100& 100 & 100  & 99.4 & \textbf{84.5} \\
        \midrule
        \multirow{3}{*}{1000} & APE & 100 & 90.8 & 79.3 & 51.8 & 14.1 \\
        & RPE$_k$ & 100 & 100 & 100 & 100 & 15.2\\
        & RPE$_{k,q}$ & 100 & 100 & 100 & 100 & 9.8\\

       \bottomrule
    \end{tabular}
    \caption{}\label{tab:modular_addition}
\end{subtable}
\hspace{.5cm}
 \begin{subtable}{.45\textwidth}
 \small
    \begin{tabular}{clccccc}
        & & & \multicolumn{4}{c}{Digits}\\
        c & PE  & 5 & 10 & 20 & 30 & 35\\ 
        \midrule
        \multirow{3}{*}{100} & \multirow{1}{*}{APE} & 100 & 98.8 & 96.2 & 90.2 & 88.1  \\
        & \multirow{1}{*}{RPE$_k$}  & 100 & 100 & 97.5 & 85.8 & 65.2 \\
        & \multirow{1}{*}{RPE$_{k,q}$}  & 100& 100 & 100 & 100 & 100 \\
        \midrule
        
        \multirow{3}{*}{1000} & \multirow{1}{*}{APE}   & 80.2 & 69.8 & 43.4 & 26.3 & 6.4 \\
        & \multirow{1}{*}{RPE$_k$}  & 100 & 84.8 & 4.9 & 0.2 & 0 \\
        & \multirow{1}{*}{RPE$_{k,q}$}  & 100 & 97.9 & 82.6 & 55.1 & 3.9\\

       \bottomrule
    \end{tabular}
    \caption{} \label{tab:modular_multiplication}

 \end{subtable}

\caption{\small\textbf{Modular addition and multiplication:} (a) Extrapolation results for addition and (b) for multiplication. We train a UTransformer in its base version ($D=6,\dmodel=512,h=8$) with three position embedding methods (APE, RPE$_k$, RPE$_{k,q}$). We report the accuracy  on 100,000 example test sets. }

\vspace{-.4cm}
    
\end{table}



\paragraph{Modular multiplication.} In the easy cases ($c\in \{100, 1000\}$), both APE and RPE-based model generalize, achieving $100\%$ on $35$-digit numbers for $c=100$. For $c=1000$, APE achieve $43\%$ on $20$-digit numbers, but the use of RPE improves performance, to $83\%$ on $20$-digit numbers and $55\%$ on $30$-digit numbers (\autoref{tab:modular_multiplication}). 
On hard instances (see \autoref{sec:appendix}), for $c=128$ , the model performance drops, both in and out of domain, but length generalization still happens, and is facilitated by RPE and larger models. Finally, for $c=101$, models can learn modular multiplication in-domain, but consistently fail on longer sequences.
Modular multiplication turns out to be easier to learn than modular addition. A possible explanation is the fact that multiplication tables display more redundancy, that the model can exploit, than addition tables.\newline

Our experiments with modular arithmetic help understand the role of position embeddings. APE-based models generalize when they learn an operation involving a fixed number of input tokens, and constant length output.


%% file: multiplication.tex
\section{Multiplication: train set priming for length generalization}\label{sec:multiplication}

We focus on the length generalization problem where we train a UTransformer to multiply $5$-digit numbers by $3$-digit numbers, from $\Ntrain=5000$ examples and train it on a set of $\Ntrain=5000$ examples that are $(\ntrain\times3)$-multiplications with $\ntrain\leq 5.$ We test its extrapolation ability to perform $35\times 3$ multiplications.


\begin{table}
    \small
    \centering
    \begin{tabular}{clcccc}
        Second & & \multicolumn{4}{c}{Digits}\\
        operand & PE & & 5 & 6 & 7 \\ 
        \midrule
        \multirow{3}{*}{1-digit} & APE & & 100 & 1.5 & 0 \\
        & RPE$_k$ & & 100 & 12.2 & 0\\
        & RPE$_{k,q}$&  & 100 & 9.2 & 0 \\
        \midrule
        \multirow{3}{*}{2-digits} & APE & & 100 & 0 & 0 \\
        & RPE$_{k}$ & & 100 & 16.9 & 0 \\
        & RPE$_{k,q}$ & &  100 & 15.5 & 0\\
        \midrule
        \multirow{3}{*}{3-digits} & APE  & &  100 & 0 & 0\\
        & RPE$_{k}$ & & 98.9 & 0 & 0\\
        & RPE$_{k,q}$ & & 100 & 0 & 0 \\

       \bottomrule
    \end{tabular}
\caption{\small\textbf{Multiplication by $1,2$ and $3$-digit numbers}: We train a UTransformer in its standard version ($D=6,\dmodel=1024,h=16$) with three position embeddings (APE, RPE$_k$, RPE$_{k,q}$). ID and OOD accuracy on 100,000 test examples.}
    \label{tab:multiplication_base}
\end{table}



\subsection{Relative position embeddings and fine-tuning}

\paragraph{Relative position embeddings are not sufficient.} We first train UTransformers with the three position embedddings (\autoref{tab:multiplication_base}). All models achieve close to $100\%$ in-domain accuracy, but fail to generalize to numbers with $6$ digits or more. For $5 \times 3$ multiplication, RPE do not generalize. On simpler versions of this task ($5\times 2$ and $5 \times 1$), RPE models achieve limited generalization to $6$-digit numbers ($12.2$ and $16.9\%$ for $1$ and $2$-digits), but fail for longer sequences. \newline

\begin{figure}[t]
\begin{subfigure}{0.49\columnwidth}
\centering
\includegraphics[width=.8\linewidth]{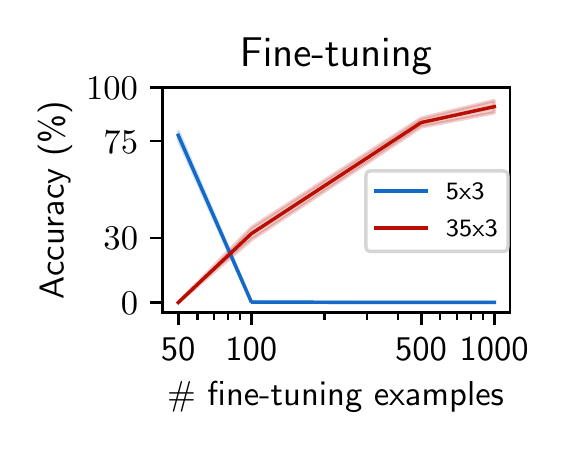}

\vspace{-.5cm}
\caption{}\label{fig:finetuning}

\end{subfigure}
\begin{subfigure}{0.49\columnwidth}

\centering
\includegraphics[width=.8\linewidth]{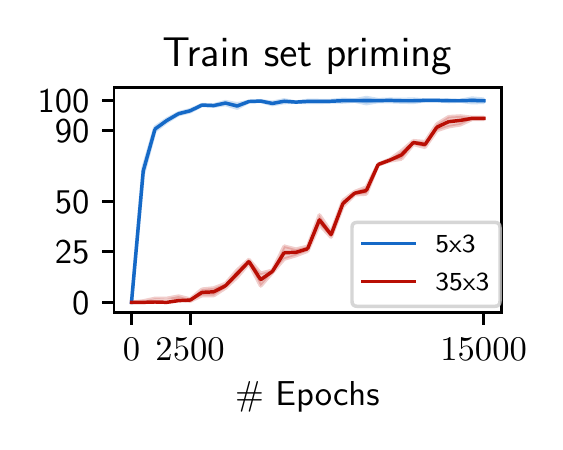}

\vspace{-.5cm}

\caption{}\label{fig:convergence_plot}
\end{subfigure}

\vspace{-.1cm}

\caption{\textbf{Fine-tuning {\normalfont (a)} and train set priming {\normalfont (b)}.} (a) fine-tuning, the model is trained on $5\times 3$ multiplications, then fine-tuned on $35\times 3$ multiplications. 
Final accuracy of $5\times 3$ and $35\times 3$ multiplications as a function of the number of fine-tuning examples.  (b) priming,  fifty $35\times 3$ examples are added to the training set. Learning curves for $5$-digit and $35$-digit accuracy. All experiments use a standard UTransformer ($D=6,\dmodel=1024,h=16$). Average over 3 seeds. }


\end{figure}

\paragraph{Fine-tuning requires a sizable sample set.} Fine-tuning is a common solution for transfer learning (extrapolating from one distribution to another). Here, we first train a model on $5 \times 3$ multiplication, then re-train it on a fixed sample of $35 \times 3$ examples. We observe (\autoref{fig:finetuning}) that $35$-digit multiplication can indeed be learned by fine-tuning on a set of $1000$ examples. This is a large number: as we shall see, train set priming allows for much smaller samples.  Besides, the fine-tuned model is not longer able to perform $5\times 3$ multiplication, a phenomenon known as catastrophic forgetting \citep{mccloskey1989catastrophic}. 

\subsection{Priming for length generalization in multiplication.} 

As an alternative, we introduce \textit{train set priming}: adding a tiny amount ($\varepsilon\%$) of long sequences to the training set. By adding $50$ $35$-digit examples ($\varepsilon=1\%$), our model achieves close to $100\%$ accuracy on $5\times 3$ and $35 \times 3$ multiplication (\autoref{fig:convergence_plot}). To reach equivalent performance, train sample priming needs $20$ times less examples than fine-tuning. $5\times 3$ multiplication is learned after a few hundred thousand examples, $35 \times 3$ multiplication (OOD generalization) after $1500$ epochs, or $7.5$ million examples ($1500$ passes over $5000$ fixed examples), but only $75,000$ $35$-digit example (i.e. $1,500$ passes over $50$ fixed examples, out of $9.10^{34}$ possible 35-digit integers).

\begin{figure}[t]
\hspace{-.8cm}
\begin{subfigure}{0.245\columnwidth}

\centering
\includegraphics[width=1.1\linewidth]{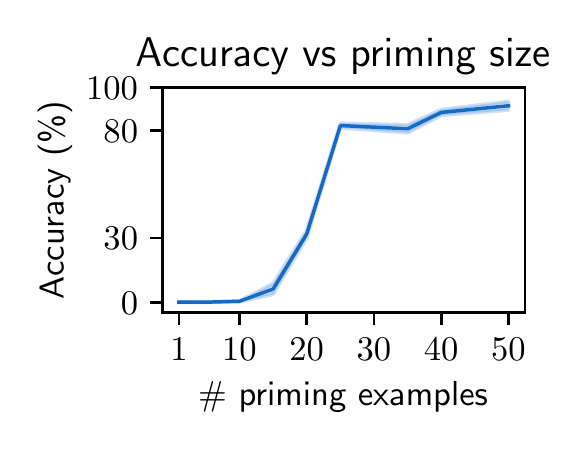}

\vspace{-.4cm}
\caption{}\label{fig:phase_transition}
\end{subfigure}
\hspace{.1cm}
\begin{subfigure}{0.245\columnwidth}

\centering

\includegraphics[width=1.1\linewidth]{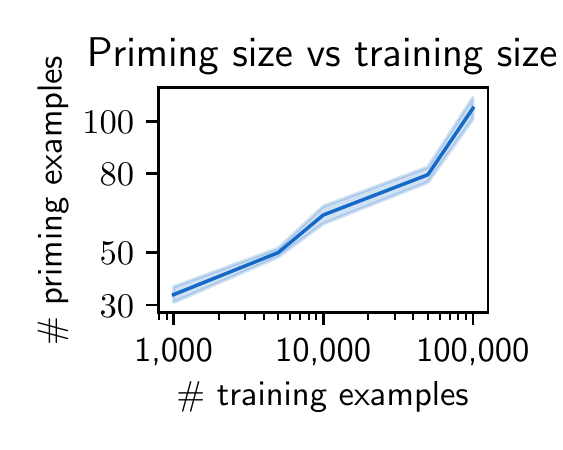}

\vspace{-.4cm}
\caption{}\label{fig:trainingsize}
\end{subfigure}
\hspace{.1cm}
\begin{subfigure}{0.245\columnwidth}

\centering

\includegraphics[width=1.1\linewidth]{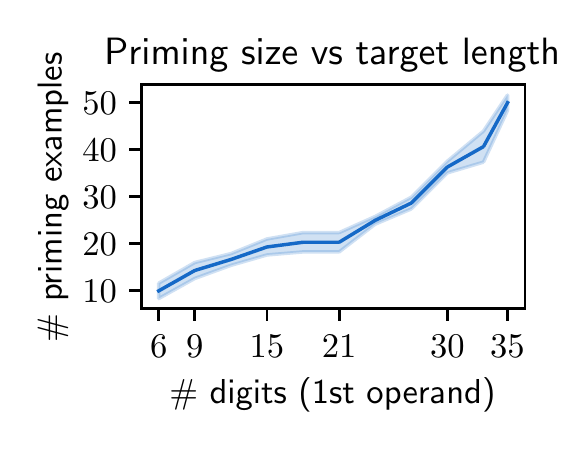}

\vspace{-.4cm}

\caption{}\label{fig:fewshotdigits}
\end{subfigure}
\hspace{.1cm}
\begin{subfigure}{0.245\columnwidth}

\centering

\includegraphics[width=1.1\linewidth]
{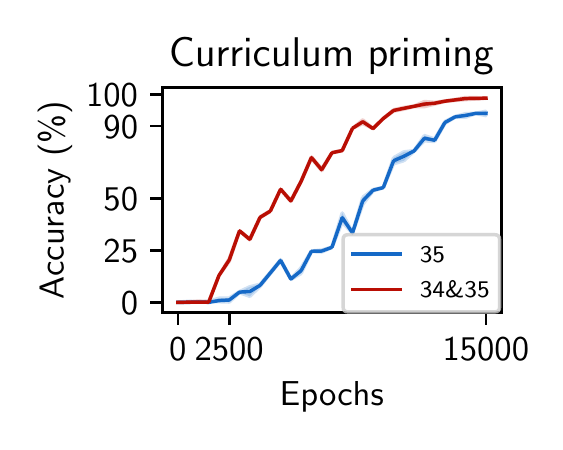}

\vspace{-.4cm}

\caption{}\label{fig:understandmul}
\end{subfigure}

\caption{\textbf{Ablations on priming sample size.} (a) Accuracy of $35\times3$-multiplications vs priming sample size. (b) Priming sample needed to achieve $90\%$ $35$-digit accuracy for different train set sizes. (c) Priming sample needed to achieve $90\%$ accuracy, for different extrapolation lengths. (d) Learning curves for $35$-digit priming, and $34$ and $35$-digit curriculum. 
All experiments use a standard UTransformer ($D=6,\dmodel=1024,h=16$). Results are averaged over 3 seeds.} 
\label{fig:ablation_fewshot}


\end{figure}

\paragraph{A minimal priming rate is required.} Adding less than $25$ samples (25 examples, $\varepsilon=0.5\%$) prevents generalization. Over that threshold, accuracy increases with the priming rate (\autoref{fig:phase_transition}). 

\paragraph{Priming sample scales logarithmically with train set size.}  As the number of training examples increases, so does the number of priming examples required to extrapolate to $35\times 3$. However, it scales \textit{logarithmically}: $30$ ($\varepsilon$=$3\%$) priming examples are needed for $10^3$ training examples, $70$ ($\varepsilon$=$0.7\%$) for $10^4$ and $100$ ($\varepsilon$=$0.1\%$) for $10^5$ (\autoref{fig:trainingsize}).

\paragraph{Priming sample scales linearly with extrapolation length.}   Whereas $50$ samples are needed for $35$-digit generalization, $6$-digit generalization only needs $10$ (\autoref{fig:fewshotdigits}). 
 

\paragraph{Curriculum priming fails.} 
We consider \textit{curriculum priming} as a possible improvement. Instead of priming on long sequences only (i.e. $35$-digit numbers), we could split the priming examples between several lengths, from $6$ to $35$. 
In most cases, curriculum priming fails to extrapolate to $35\times3$  multiplication, but one curriculum proves effective: priming the model on a mixture of $34$ and $35$-digits numbers (\autoref{fig:understandmul}). This causes the model 
to learn faster and achieve higher extrapolation accuracy.

\subsection{Priming for extrapolation at all lengths}

\begin{figure}[t]
\begin{subfigure}{0.245\columnwidth}
 \hspace{1cm}
 \includegraphics[width=1.1\linewidth]{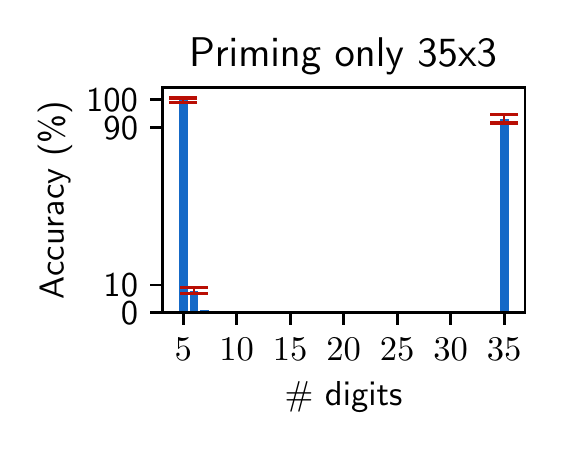}
 
 \vspace*{-.4cm}
 
 \caption{}\label{fig:itd_vanilla}
\end{subfigure}
\begin{subfigure}{0.245\columnwidth}
\hspace{1cm}
 \includegraphics[width=1.1\linewidth]{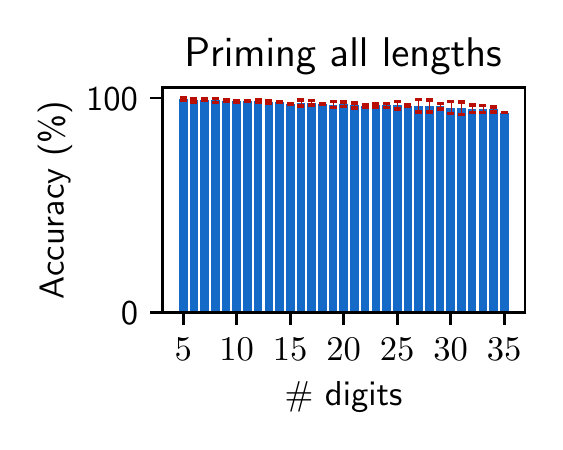}
 
 \vspace*{-.4cm}
 
 \caption{}\label{fig:hist_fewshot}
\end{subfigure}
\begin{subfigure}{0.245\columnwidth}
\hspace{1cm}
 \includegraphics[width=1.1\linewidth]{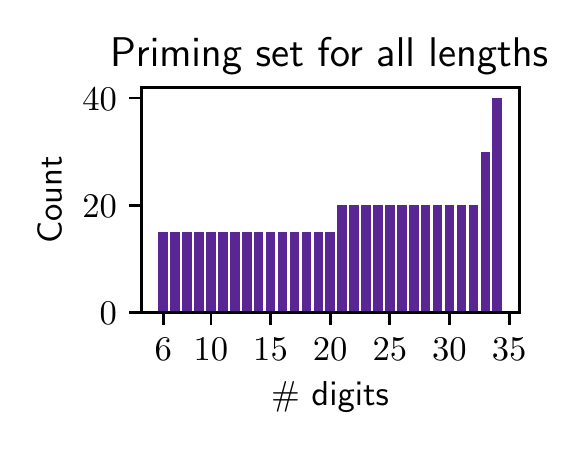}
 
 \vspace*{-.4cm}
 
 \caption{}\label{fig:hist_data}
\end{subfigure}
\begin{subfigure}{0.245\columnwidth}
\hspace{1cm}
\includegraphics[width=1.1\linewidth]{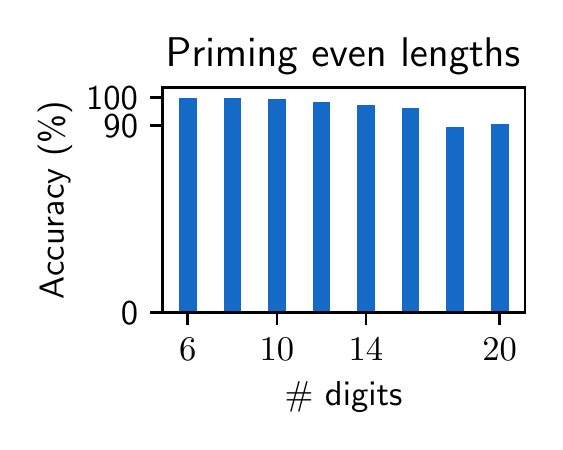}

 \vspace*{-.4cm}
\caption{}\label{fig:holes}
\end{subfigure}

\vspace{-3mm}
\caption{\small \textbf{Training set priming to all lengths.} (a) Priming with $35$-digit numbers only. (b) Priming with a mixture of all length. (c) Distribution of priming lengths for figure (b). (d) Priming on even lengths only. 
All experiments use a standard UTransformer ($D$ = $6$, $\dmodel$ = $1024$, $h$ =$16$). Average over 3 seeds. }


\end{figure}

Priming the train set with $35$-digit numbers only allows to  extrapolate to $35$-digit operands. No other extrapolation lengths are learned in the process (\autoref{fig:itd_vanilla}).
However, by priming on numbers of \textit{all} lengths from $6$ to $35$, the model can extrapolate to all lengths up to $35$.
This can be done at a moderate cost in additional data. 
Using the priming distribution from \autoref{fig:hist_data}, our models learn to extrapolate with over $95\%$ accuracy to all lengths (see Figure~\ref{fig:hist_fewshot}). The priming set size is $500$, for a priming rate of $\varepsilon=10\%$. More efficient priming distributions might exist: the point of this experiment is to show that priming to all lengths is possible within a reasonable data budget $\varepsilon.$ 
  On the other hand, we observe that all extrapolation length must be primed. For instance, if only even lengths are primed, the model only generalizes to even lengths. There is no overspill to odd lengths (\autoref{fig:holes}).

%% file: discussion.tex
\section{Discussion}\label{sec:discussion}

\subsection{Why do RPEs extrapolate better than APEs?} \label{sec:aperep}

In \autoref{sec:addition}, we notice that replacing APE by RPE is the key for models to length generalize. Three experiments help understand the role of RPE. \newline

\textbf{Element-wise addition.} A possible reason for generalization in RPE-based models, is that relative embeddings allow tokens to ``know their neighbors''. This could help models learn local operations, like carry propagation (an important factor in integer addition). To test this hypothesis, we train models on element-wise addition $\oplus$ (i.e. addition without carries: $99\oplus35=24$). If carry propagation is the reason why RPE succeed, APE-models should generalize on this task. Experimental results (in \autoref{sec:appendix}) show that APE fail to generalize on element-wise addition, whereas RPE succeed, this disproving our hypothesis. It is  striking to note (see \autoref{fig:ewise}) that when the generalize, APE models almost always predict the the $5$ leftmost digits of the results, i.e. its ``in-domain'' positions, thus confirming our intuition that APE learn addition digit by digit.\newline


\textbf{Modular arithmetic.} As we have seen, APE models length generalize on these tasks when the modulus is a power of 10. (Tables~\ref{tab:modular_addition} and~\ref{tab:modular_multiplication}). In both cases, the model output have constant length. This, together with our element-wise results, suggest that varying output lengths are an important factor of APE extrapolation failures.\newline

\begin{figure*}[t]
\begin{subfigure}{0.245\textwidth}
 \hspace{1cm}
 \includegraphics[width=.99\linewidth]{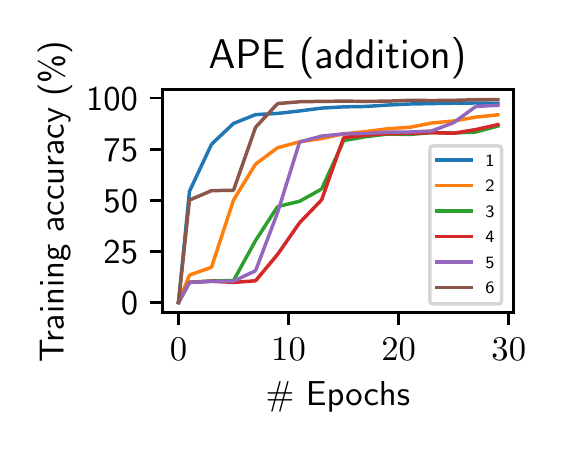}
 
 \vspace*{-.4cm}
 
 \caption{}\label{fig:add_ape}
\end{subfigure}
\begin{subfigure}{0.245\textwidth}
\hspace{1cm}
 \includegraphics[width=.99\linewidth]{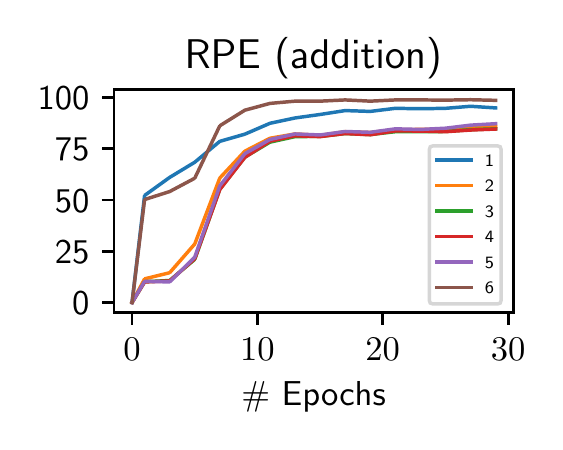}
 
 \vspace*{-.4cm}
 
 \caption{}\label{fig:add_rpe}
\end{subfigure}
\begin{subfigure}{0.245\textwidth}
 \hspace{1cm}
 \includegraphics[width=.99\linewidth]{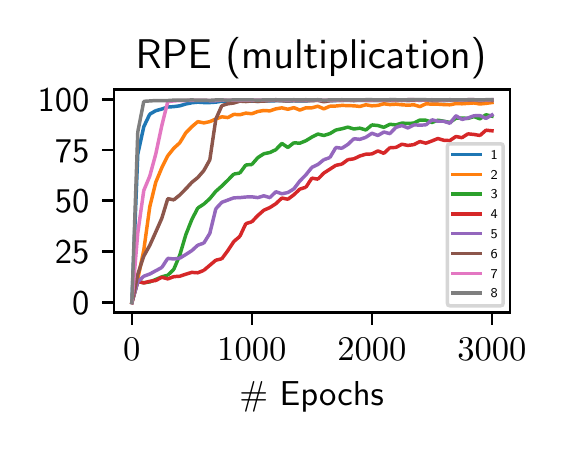}
 
 \vspace*{-.4cm}
 
 \caption{}\label{fig:mul_rpe}
\end{subfigure}
\begin{subfigure}{0.245\textwidth}
\hspace{1cm}
 \includegraphics[width=.99\linewidth]{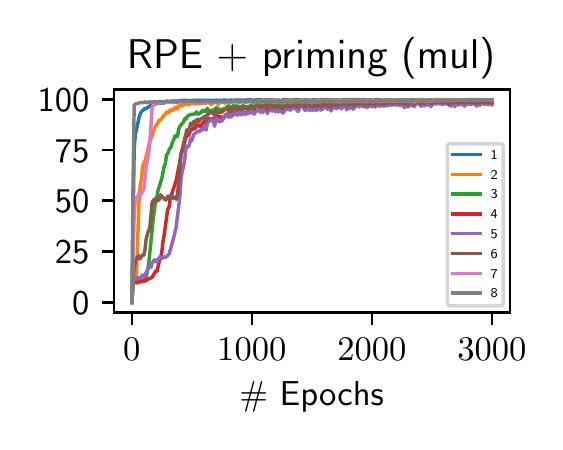}
 
 \vspace*{-.4cm}
 
 \caption{}\label{fig:mul_fewshot}
\end{subfigure}

\vspace{-3mm}
\caption{\small \textbf{Digit by digit learning curves.} Training accuracy for each output digit ($1$ are the largest powers, $6$ the units for a sum).(a) Addition APE models. (b) Addition RPE models. (c) Multiplication RPE models (no priming) (d). Multiplication RPE models (with priming). In all these experiments, $1$ denotes the leftmost digit position while $6$ (for addition) and $8$ (for multiplication) All experiments use a standard UTransformer ($D=6,\dmodel=1024,h=16$). }
\vspace{-.4cm}

\end{figure*}

\textbf{RPE-models learn all digits at once.} Figures \ref{fig:add_ape} and \ref{fig:add_rpe} present learning curves for each position in the output, when a model is trained on $5$-digit addition (e.g. the $6$ curve is the learning curve of the units of the sum, the $5$-curve is the tens). We note that whereas the first and last digits in the sums are learned first, all other digits are learned simultaneously by RPE models, whereas APE models seem to learn each position independently. This suggests that RPE models might learn a single algorithm for all positions, which greatly helps them to generalize.

\subsection{Failure cases in addition} 
\begin{figure*}[t]
\begin{subfigure}{0.245\textwidth}
 \hspace{1cm}
 \includegraphics[width=.99\linewidth]{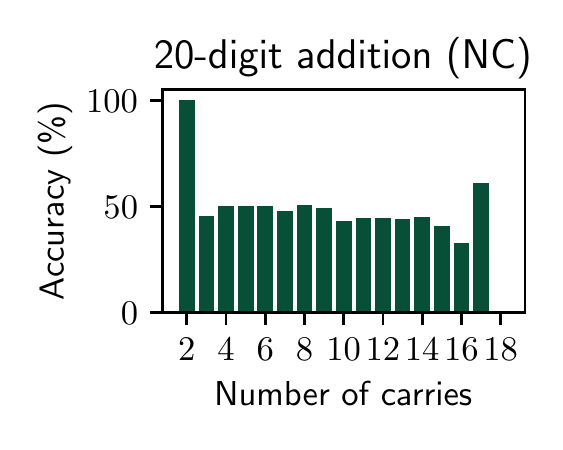}
 
 \vspace*{-.4cm}
 
 \caption{}\label{fig:add_nc}
\end{subfigure}
\begin{subfigure}{0.245\textwidth}
\hspace{1cm}
 \includegraphics[width=.99\linewidth]{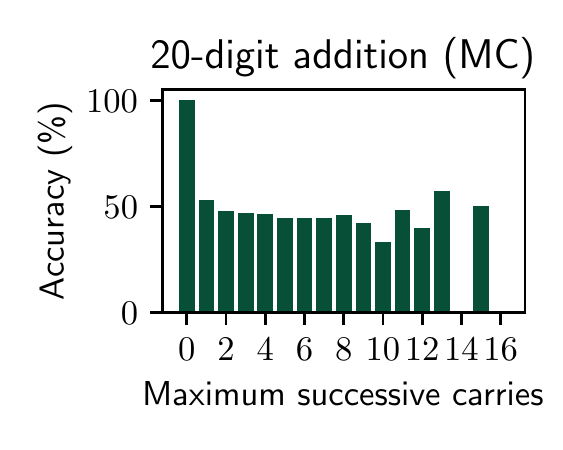}
 
 \vspace*{-.4cm}
 
 \caption{}\label{fig:add_mc}
\end{subfigure}
\begin{subfigure}{0.245\textwidth}
 \hspace{1cm}
 \includegraphics[width=.99\linewidth]{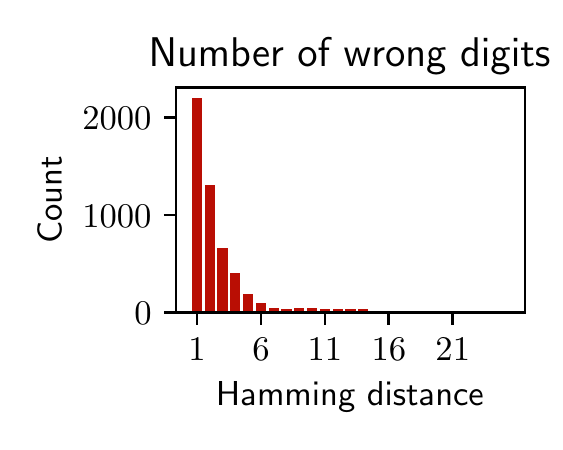}
 
 \vspace*{-.4cm}
 
 \caption{}\label{fig:ham_}
\end{subfigure}
\begin{subfigure}{0.245\textwidth}
\hspace{1cm}
 \includegraphics[width=.99\linewidth]{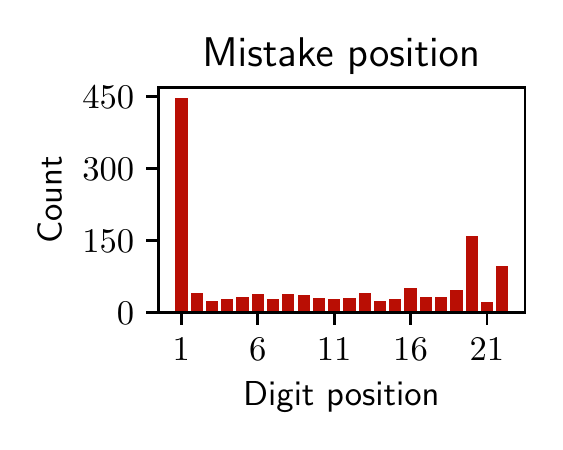}
 
 \vspace*{-.4cm}
 
 \caption{}\label{fig:digitpositionham1}
\end{subfigure}

\vspace{-3mm}
\caption{\small\textbf{Success and failure cases in addition}. (a) Accuracy of $20$-digit sums, by number of carries in the sum. (b) Accuracy of $20$-digit sums, by maximum number of consecutive carries. (c) 
Distribution of the number of incorrect digits in wrong predictions of $20$-digit sums. (d) Positions of incorrect digits in sumes where only one digit is wrong. 
All experiments use a standard UTransformer ($D=6,\dmodel=1024,h=16$), achieving $57\%$ accuracy on $20$-digit additions.} 
\label{fig:add_failure}
\end{figure*}

\autoref{fig:add_failure} provides an analysis of model failures when extrapolating to $20$-digit sums. First, we assess the role of carries, by introducing two metrics: the total number of carries (NC), and the maximum number of consecutive carries (MC). As Figures \ref{fig:add_nc} and \ref{fig:add_mc} indicate, almost all model failures happen on additions involving at least three carries, and two consecutive carries. Larger values of MC and NC have no further impact.\newline

Figures~\ref{fig:ham_} and~\ref{fig:digitpositionham1} present the number of incorrect digits in wrong model predictions and their position. We note that, when wrong, the model usually does not hallucinate a irrelevant answer (with many wrong digits), but fails on just a few. Errors also concentrate on the first and second positions: the largest powers of ten in the sum.

\subsection{More about priming}

Train set priming is our most striking result. In \autoref{sec:multiplication}, we demonstrate that is allows length generalization in multiplication. We now present additional results. We first show that train set priming is also effective on APE models. Then, we investigate how the models learn multiplication. 

\paragraph{Primed APE models generalize.} In \autoref{sec:appendix}, we show that priming on APE models also yields length generalization. We obtain a similar dynamics as in \autoref{fig:convergence_plot} where the ID accuracy quickly increases and the OOD accuracy slowly follows (\autoref{fig:ape_accuracy}). However, as expected, this does not make APE models a viable proposition: the priming rate needed is $10$ times larger i.e.\ $\varepsilon=10\%$.

\paragraph{Primed models learn several digits simultaneously.} In our addition experiments in \autoref{sec:aperep}, we noticed that whereas APE models learn to predict their output digit by digit as training proceeds (\autoref{fig:add_ape}), RPE models seem to learn them all at once (\autoref{fig:add_ape}). A similar pattern can be seen for multiplication with RPE models. Without priming (\autoref{fig:mul_rpe}), models seem to learn $5\times 3$ multiplication one digit at a time, over $1000$ epochs. With priming, the model seems to learns several digits concurrently \autoref{fig:mul_fewshot}. A similar phenomenon holds for APE models: without priming, the model independently learns each digit 
(\autoref{fig:digits_ape}) while the digits are concurrently learnt with priming (\autoref{fig:ape_digits_primig}). In summary, simultaneous learning of all the training digit positions seems a key determinant of length generalization. 


\subsection{Priming beyond arithmetic} 
Our work demonstrates that train set priming can improve the length generalization of transformers on arithmetic tasks. Compared to fine-tuning, it requires much fewer samples from the target distribution and allows for generalization without catastrophic forgetting.
We conclude on a number of open questions, which constitute as many avenue for future research. All these directions may help shed light on the capabilities and limitations of transformers, and inspire new methods for improving their generalization and adaptation. 
\begin{itemize}
   
   \item[--]\textbf{Can priming be extended to other mathematical problems?} For instance, numerical computations, matrix operations, or symbolic mathematics.
   \item[--]  
   
   \textbf{Can priming help with compositionality?} Investigate the limits of length generalization in terms of the number and type of operations. For instance, if we train on adding $k$ numbers, can we generalize to adding $k+1$ numbers, or if we train on compositions of additions and multiplications separately, does it generalize to compose them together? 
 
   \item[--] \textbf{Theoretical understanding of priming}: why is train set priming more effective than fine-tuning for length generalization? 
   \item[--] 
   
   \textbf{Can priming work for NLP?} Can we use priming to adapt a pre-trained language model to a new language task, without losing its performance on the original data? 
\end{itemize}

%% file: appendix.tex
\section{Additional experiments}\label{sec:appendix}

In this section, we present some additional experiments that were mentioned in the paper. We first provide in \autoref{app:modarithmetic} the complete results for modular addition and  multiplication that were mentioned in \autoref{sec:modular}. We then present complementary results to our discussion in \autoref{sec:discussion}. We first report the results obtained by APE and RPE models on digitwise addition. Then, we show that APE models can also be primed to length generalize in multiplication at the expense of a much larger priming rate (\autoref{sec:ape_mul}). Lastly, we present plots showing the digit order by which RPE and APE models make the correct predictions (\autoref{app:prediction}).

\subsection{Additional experiments on modular arithmetic}\label{app:modarithmetic}


\begin{table}[h]
    \small
    \centering
    \begin{tabular}{cllccccc}
        & & & \multicolumn{4}{c}{Digits}\\
        c & PE & Size & 5 & 10 & 20 & 30 & 35\\ 
        \midrule
        \multirow{6}{*}{100} & \multirow{2}{*}{APE} & Base & 100 & 98.8 & 96.2 & 90.2 & 88.1  \\
        & & Large & 100 & 100 & 100 & 100 & 100   \\
        & \multirow{2}{*}{RPE$_k$} & Base & 100 & 100 & 97.5 & 85.8 & 65.2 \\
        & & Large & 100&  100 & 100 & 100 & 100 \\
        & \multirow{2}{*}{RPE$_{k,q}$} & Base & 100& 100 & 100 & 100 & 100 \\
        && Large & 100 &  100 & 100 & 100 & 100  \\        
        \midrule
        \multirow{6}{*}{1000} & \multirow{2}{*}{APE} & Base & 80.2 & 69.8 & 43.4 & 26.3 & 6.4 \\
        & & Large & 28.2 & 12.2 &9.9 & 8.7 & 7.7 \\
        & \multirow{2}{*}{RPE$_k$} & Base & 100 & 84.8 & 4.9 & 0.2 & 0 \\
        & & Large & 100 & 100 & 100 & 99.9 & 26.4 \\
        & \multirow{2}{*}{RPE$_{k,q}$} & Base & 100 & 97.9 & 82.6 & 55.1 & 3.9\\
        && Large & 100 & 84.2 & 83.0 & 82.7 & 20.1  \\
        \midrule
        \multirow{6}{*}{128} & \multirow{2}{*}{APE} & Base & 14.7 & 8.4 & 4.7 & 4.4 & 3.8 \\
        & & Large & 9.1 &  6.9 & 5.3 & 4.4 & 3.9 \\
        & \multirow{2}{*}{RPE$_k$} & Base & 19.9 & 13.3 & 5.6 & 3.5 & 1.2 \\
        & & Large & 11.8 & 11.5 & 11.4 & 11.2 & 10.0 \\
        & \multirow{2}{*}{RPE$_{k,q}$} & Base & 26.9 & 21.7 & 14.1 & 10.3 & 6.2\\
        && Large & 20.4 & 20.5 & 19.2 & 18.4 & 16.2  \\
        \midrule

        \multirow{6}{*}{101} & \multirow{2}{*}{APE} & Base & 44.8 & 2.3 & 2.4 & 2.4 & 2.3\\
        & & Large & 1.1 & 1.2 & 1.2 & 1.1 & 1.1\\
        & \multirow{2}{*}{RPE$_k$} & Base & 24.5 & 2.3 & 1.9 & 1.8 & 1.4 \\
        & & Large & 95.3 & 2.3  & 2.2 & 2.0 & 2.1 \\
        & \multirow{2}{*}{RPE$_{k,q}$} & Base & 99.1 & 2.5 & 2.2 & 2.2 & 2.1 \\
        && Large & 9.9 & 2.4 & 2.1 & 1.8 & 1.8  \\

       \bottomrule
    \end{tabular}
\caption{\small\textbf{Modular addition:} Extrapolation results for modulo $c\in \{100, 1000, 128,101\}$. UTransformer model in their Base and Large format.  We report the accuracy reached by the models on 100,000 example test sets.  } 
    \label{tab:modular_addition_final}
\end{table}

\autoref{tab:modular_addition_final} provides a more complete version of \autoref{tab:modular_addition} where we do modular addition  for modulus $c\in\{128,101\}$. As explained in \autoref{sec:modular}, the model manages to extrapolate when the modulus is a power of $10$. When $c=128,101$, the model  fails to extrapolate. This shows that what the model struggles when the length of the digits that matter vary.



\begin{table}[h]
    \small
    \centering
    \begin{tabular}{cllccccc}
        & & & \multicolumn{4}{c}{Digits}\\
        c & PE & Size & 5 & 10 & 20 & 30 & 35\\ 
        \midrule
        \multirow{6}{*}{100} & \multirow{2}{*}{APE} & Base & 100 & 98.8 & 96.2 & 90.2 & 88.1  \\
        & & Large & 100 & 100 & 100 & 100 & 100   \\
        & \multirow{2}{*}{RPE$_k$} & Base & 100 & 100 & 97.5 & 85.8 & 65.2 \\
        & & Large & 100&  100 & 100 & 100 & 100 \\
        & \multirow{2}{*}{RPE$_{k,q}$} & Base & 100& 100 & 100 & 100 & 100 \\
        && Large & 100 &  100 & 100 & 100 & 100  \\        
        \midrule
        \multirow{6}{*}{1000} & \multirow{2}{*}{APE} & Base & 80.2 & 69.8 & 43.4 & 26.3 & 6.4 \\
        & & Large & 28.2 & 12.2 &9.9 & 8.7 & 7.7 \\
        & \multirow{2}{*}{RPE$_k$} & Base & 100 & 84.8 & 4.9 & 0.2 & 0 \\
        & & Large & 100 & 100 & 100 & 99.9 & 26.4 \\
        & \multirow{2}{*}{RPE$_{k,q}$} & Base & 100 & 97.9 & 82.6 & 55.1 & 3.9\\
        && Large & 100 & 84.2 & 83.0 & 82.7 & 20.1  \\
        \midrule
        \multirow{6}{*}{128} & \multirow{2}{*}{APE} & Base & 14.7 & 8.4 & 4.7 & 4.4 & 3.8 \\
        & & Large & 9.1 &  6.9 & 5.3 & 4.4 & 3.9 \\
        & \multirow{2}{*}{RPE$_k$} & Base & 19.9 & 13.3 & 5.6 & 3.5 & 1.2 \\
        & & Large & 11.8 & 11.5 & 11.4 & 11.2 & 10.0 \\
        & \multirow{2}{*}{RPE$_{k,q}$} & Base & 26.9 & 21.7 & 14.1 & 10.3 & 6.2\\
        && Large & 20.4 & 20.5 & 19.2 & 18.4 & 16.2  \\
        \midrule

        \multirow{6}{*}{101} & \multirow{2}{*}{APE} & Base & 44.8 & 2.3 & 2.4 & 2.4 & 2.3\\
        & & Large & 1.1 & 1.2 & 1.2 & 1.1 & 1.1\\
        & \multirow{2}{*}{RPE$_k$} & Base & 24.5 & 2.3 & 1.9 & 1.8 & 1.4 \\
        & & Large & 95.3 & 2.3  & 2.2 & 2.0 & 2.1 \\
        & \multirow{2}{*}{RPE$_{k,q}$} & Base & 99.1 & 2.5 & 2.2 & 2.2 & 2.1 \\
        && Large & 9.9 & 2.4 & 2.1 & 1.8 & 1.8  \\

       \bottomrule
    \end{tabular}
\caption{\small\textbf{Modular multiplication:} Extrapolation results for modulo $c\in \{100, 1000, 128,101\}$. UTransformer model in their Base and Large format.  We report the accuracy reached by the models on 100,000 example test sets.  } 
    \label{tab:modular_multiplication_final}
\end{table}

\autoref{tab:modular_multiplication_final} provides a more complete version of \autoref{tab:modular_multiplication} where we do modular multiplication for modulus $c\in\{128,101\}$. As explained in \autoref{sec:modular}, the model manages to extrapolate when the modulus is a power of $10$. When $c=128$, the model non-trivially length generalize while when $c=101$, the model fails to extrapolate. We do not fully know why this difference happens but one hypothesis is that $101$ is a prime number while $128$ a power of $2.$ 

\newpage

\subsection{Element-wise addition experiments}

\begin{table}[ht]
\begin{minipage}[b]{0.56\linewidth}
    \small
    \centering
    \begin{tabular}{lccccc}
         & & \multicolumn{4}{c}{Digits}\\
         PE  & 5 & 6 & 10 & 15 & 20\\ 
        \midrule
       \multirow{1}{*}{APE} & 100 & 5.3 & 0.0 & 0.0 & 0.0    \\
       \multirow{1}{*}{RPE$_k$}  & 100  & 97.5 & 90.5 & 86.2 & 78.13 \\
       
       \bottomrule
    \end{tabular}
\caption{\small\textbf{Element-wise addition}: Extrapolation results. We train a UTransformer in its base version ($D=6,\dmodel=512,h=8$) with two position embedding methods (APE, RPE$_k$). We report the accuracy reached by the models on 10,000 example test sets. } 
    \label{tab:ewise}
\end{minipage}\hfill
\begin{minipage}[b]{0.4\linewidth}

\vspace{-.5cm}

\includegraphics[width=.85\linewidth]{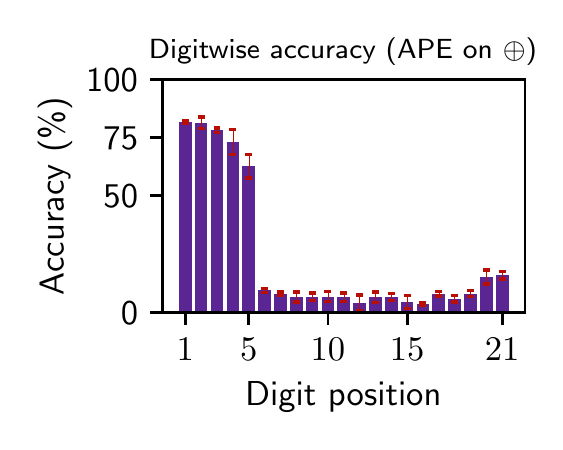}
    \captionof{figure}{\small \textbf{Digitwise accuracy of the APE model on elementwise addition.}  We train a  Base UTransformer with APEs and report the accuracy on 10,000 example test sets. Average over 3 seeds.} 
    \label{fig:ewise}
\end{minipage}
\end{table}

We consider here an element-wise  addition operation. For example, $99\oplus 45=34$ because $(9+5)\%10=4$ and $(9+4)\%10=3$. We train a UTransformer on $5$-digit element-wise addition ($\Ntrain=50,000$) and evaluate its extrapolation on $20$-digit ($\Ntest=10,000$). \autoref{tab:ewise} reports the final results obtained with APE and RPE models. We observe that the RPE models manage to length generalization while the APE models fail. In Figure \ref{fig:ewise}, we plot the digitwise accuracy on the test samples. We observe that the model managed to well-predict the leftmost 5 digits (those seen during training) but fails in the right-most ones.  

\subsection{Multiplication experiments using APEs}\label{sec:ape_mul}

\begin{figure}[h]
\begin{subfigure}{0.33\columnwidth}
 \hspace{-.6cm}
 \includegraphics[width=1.1\linewidth]{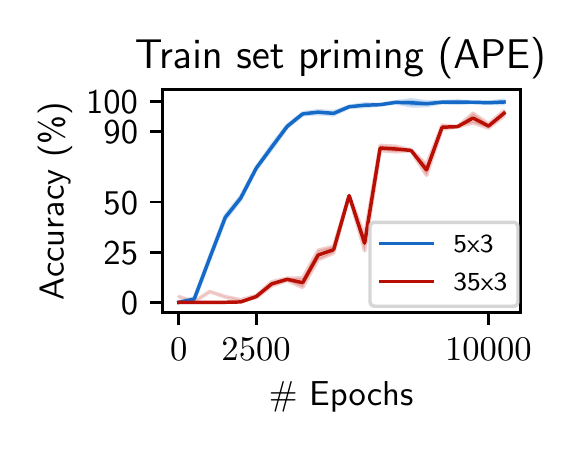}
 
 \vspace*{-.4cm}
 
 \caption{}\label{fig:ape_accuracy}
\end{subfigure}
\begin{subfigure}{0.33\columnwidth}
\hspace{-.3cm}
 \includegraphics[width=1.1\linewidth]{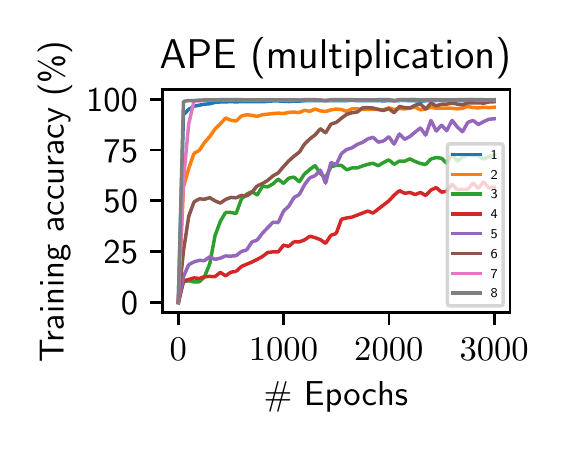}
 
 \vspace*{-.4cm}
 
 \caption{}\label{fig:digits_ape}
\end{subfigure}
\begin{subfigure}{0.33\columnwidth}
 \includegraphics[width=1.1\linewidth]{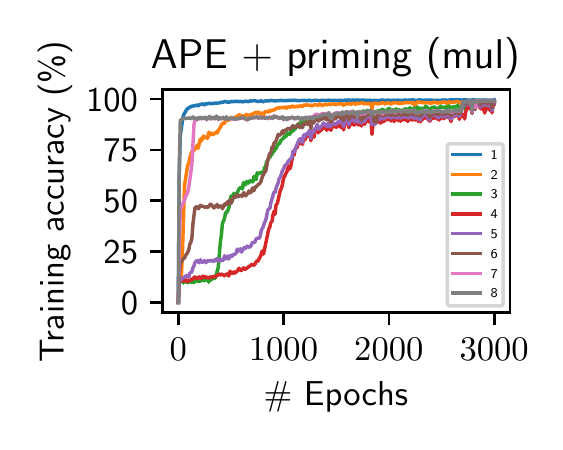}
 
\vspace*{-.4cm}
 
 \caption{}\label{fig:ape_digits_primig}
\end{subfigure}
 
 

\vspace{-3mm}
\caption{\small\textbf{Additional experiments on priming for multiplication}. (a) shows the accuracy on $5\times 3$ and $35\times 3$ multiplications obtained by an APE model. (b) and (c) respectively display the learning process of an APE model without and with train set priming on multiplication. We train a standard UTransformer ($D=6,\dmodel=1024,h=16$) on $5\times 3$-multiplications and test on $35\times 3$. Training set size is $\Ntrain=5000$ and test set size is $\Ntest=10000.$ }
\label{fig:apemodels}
\end{figure}

In this section, we consider the multiplication task with UTransformers using APEs. 
Similarly to the RPE case, we observe that training set priming lead to successful extrapolation to $(35\times 3)$-multiplications with $95\%$ test accuracy (\autoref{fig:ape_accuracy}). In \autoref{fig:digits_ape}, we observe that the model learns each digit position independently. This is a sign of memorization. On the other hand, when priming the model with $\varepsilon=10\%$, we observe that this forces the model to learn the digit positions together. A similar observation holds for RPE models in \autoref{sec:discussion}.

\subsection{Training accuracy for varying priming rates}\label{app:training}

\begin{figure}[h]
\begin{subfigure}{0.48\columnwidth}
 \hspace{-.6cm}
 \includegraphics[width=.8\linewidth]{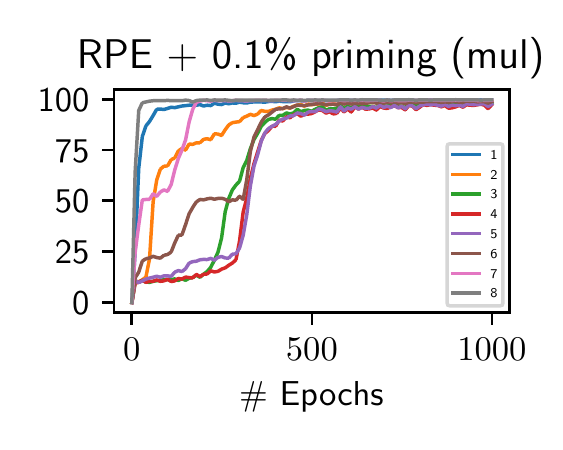}
 
 \vspace*{-.4cm}
 
 \caption{}\label{fig:priming5}
\end{subfigure}
\begin{subfigure}{0.48\columnwidth}
\hspace{.5cm}
 \includegraphics[width=.8\linewidth]{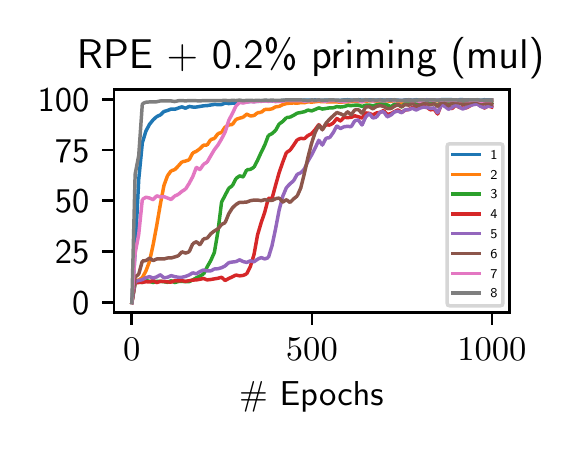}
 
 \vspace*{-.4cm}
 
 \caption{}\label{fig:priming10}
\end{subfigure}
\begin{subfigure}{0.48\columnwidth}
 \hspace{-.6cm}
 \includegraphics[width=.8\linewidth]{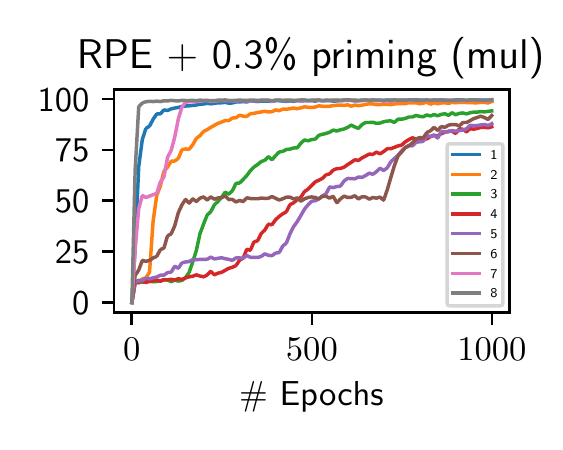}
 
 \vspace*{-.4cm}
 
 \caption{}\label{fig:priming15}
\end{subfigure}
\begin{subfigure}{0.48\columnwidth}
\hspace{.5cm}
 \includegraphics[width=.8\linewidth]{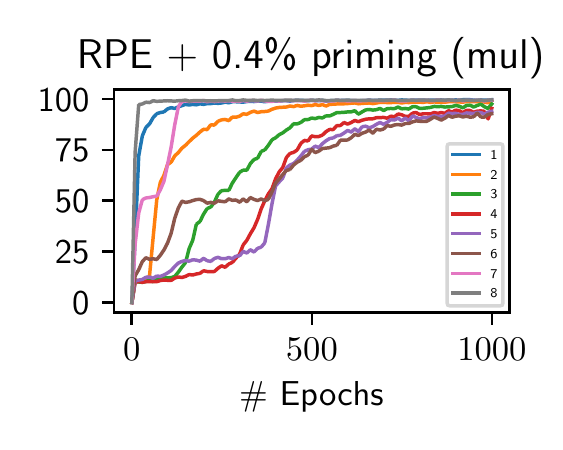}
 
 \vspace*{-.4cm}
 
 \caption{}\label{fig:priming20}
\end{subfigure}
\begin{subfigure}{0.48\columnwidth}
 \hspace{-.6cm}
 \includegraphics[width=.8\linewidth]{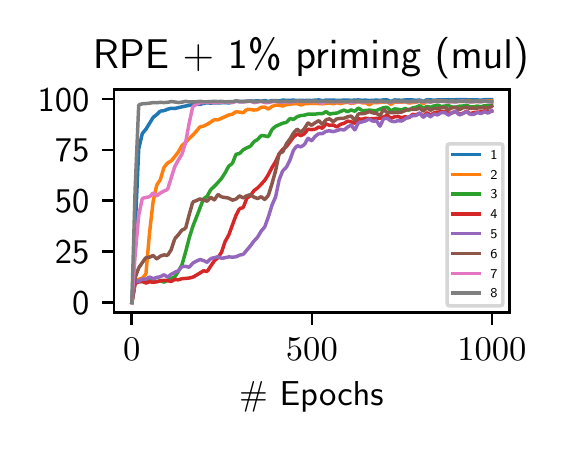}
 
 \vspace*{-.4cm}
 
 \caption{}\label{fig:priming50}
\end{subfigure}
 
 

\vspace{-3mm}
\caption{\small\textbf{Digitwise accuracy on the training examples}.  }
\label{fig:training_digitwise}
\end{figure}

\subsection{Test accuracy for varying priming rates}\label{app:prediction}

\begin{figure}[h]
\begin{subfigure}{0.48\columnwidth}
 \includegraphics[width=1.\linewidth]{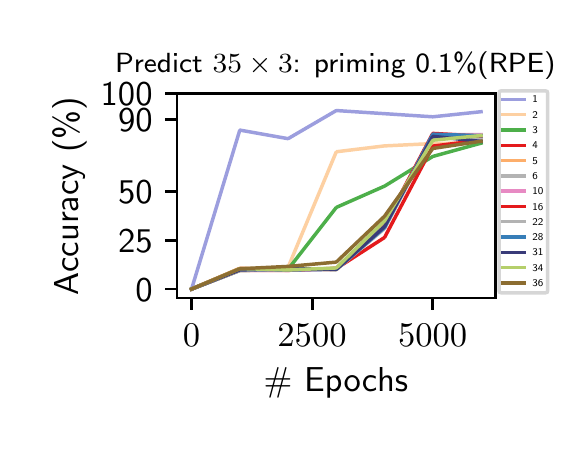}
 
 \vspace*{-.4cm}
 
 \caption{}\label{fig:priming5test}
\end{subfigure}
\begin{subfigure}{0.48\columnwidth}
 \includegraphics[width=1.\linewidth]{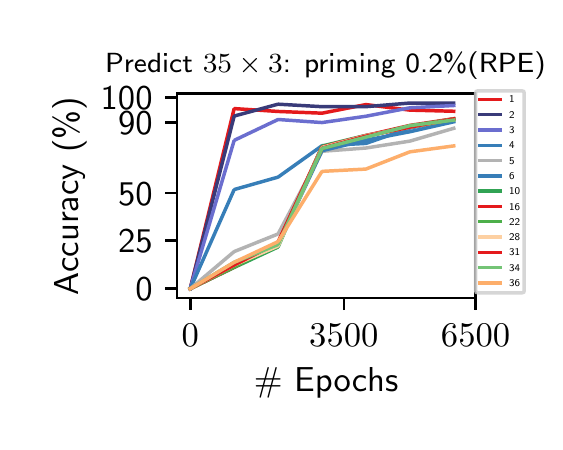}
 
 \vspace*{-.4cm}
 
 \caption{}\label{fig:primingtest10}
\end{subfigure}
\begin{subfigure}{0.48\columnwidth}
 \includegraphics[width=1.\linewidth]{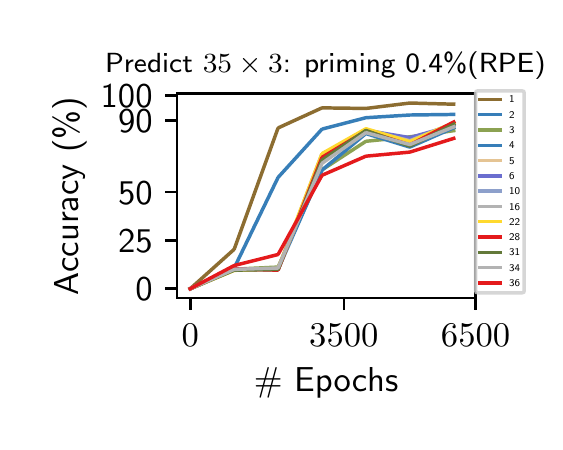}
 
 \vspace*{-.4cm}
 
 \caption{}\label{fig:priming20test}
\end{subfigure}
\begin{subfigure}{0.48\columnwidth}
 \includegraphics[width=1.\linewidth]{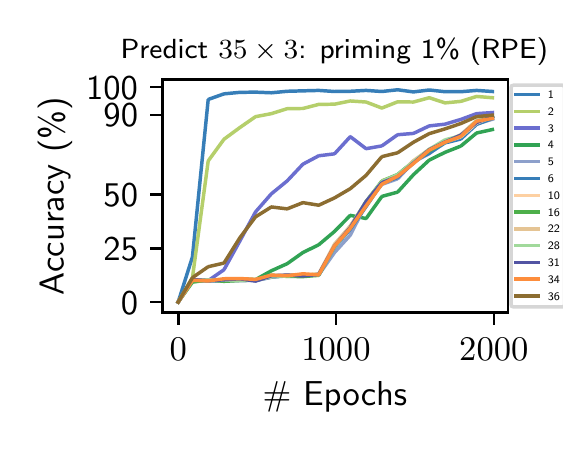}
 
 \vspace*{-.4cm}
 
 \caption{}\label{fig:priming50_test}
\end{subfigure}
 
 

\vspace{-3mm}
\caption{\small\textbf{Digitwise prediction on the test examples}.  }
\label{fig:test_digitwise}
\end{figure}